\documentclass{new_tlp}
\usepackage{mathptmx}

\title[Theory and Practice of Logic Programming]{Explanation Generation for Multi-Modal Multi-Agent Path Finding with Optimal Resource Utilization \\ using Answer Set Programming\thanks{This work is supported by Tubitak Grant~118E931.}}
\author[Bogatarkan and Erdem]{AYSU BOGATARKAN and ESRA ERDEM\\
Sabanci University, Faculty of Engineering and Natural Sciences, Istanbul, Turkey\\
\email{\{aysubogatarkan,esraerdem\}@sabanciuniv.edu}
}

\usepackage[ruled,vlined]{algorithm2e}
\usepackage{graphicx}
\usepackage{float}

\usepackage{times}
\usepackage{helvet}
\usepackage{courier}
\usepackage{xspace,caption,url}

\def\mapf{{MAPF}\xspace}
\def\mmapf{{mMAPF}\xspace}

\def\clingo{{\sc Clingo}\xspace}

\def\ba{\begin{array}}
\def\ea{\end{array}}
\def\beq{\begin{equation}}
\def\eeq#1{\label{#1}\end{equation}}

\def\ii#1{\hbox{\it #1\/}}
\def\seq#1{\left\langle #1 \right\rangle}

\def\eqs{\,{=}\,}

\def\leqs{\,{\leq}\,}

\def\lts{\,{<}\,}

\def\subseteqs{\,{\subseteq}\,}

\begin{document}

\maketitle

\begin{abstract}
The multi-agent path finding (\mapf) problem is a combinatorial search problem that aims at finding paths for multiple agents (e.g., robots) in an environment (e.g., an autonomous warehouse) such that no two agents collide with each other, and subject to some constraints on the lengths of paths. We consider a general version of \mapf, called  \mmapf, that involves multi-modal transportation modes (e.g., due to velocity constraints) and consumption of different types of resources (e.g., batteries). The real-world applications of \mmapf require flexibility (e.g., solving variations of \mmapf) as well as explainability. Our earlier studies on \mmapf have focused on the former challenge of flexibility. In this study, we focus on the latter challenge of explainability, and introduce a method for generating explanations for queries regarding the feasibility and optimality of solutions, the nonexistence of solutions, and the observations about solutions. Our method is based on answer set programming. This paper is under consideration for acceptance in TPLP.
\end{abstract}
%
%\begin{keywords}
%	
%\end{keywords}

%----------------------------------------------------

\section{Introduction}

Artificial Intelligence (AI) applications are used widely by people with different background and interests. For the success of these applications, two of the important features (and challenges) necessitated by AI methods are flexibility and explainability. A flexible AI method developed to solve a problem can accommodate variations of the problem, and thus can be used to investigate different options by people for a better understanding. An explainable AI method can provide answers to queries about the (in)feasibility and the optimality of solutions.  One of the well-studied problems in AI that necessitates solutions for these two challenges is the multi-agent path finding (\mapf) problem.

\mapf problem aims to find plans for multiple agents in an environment without colliding with each other or obstacles. Optimal solutions can be found by optimizing the total plan length of agents or the makespan of the whole plan. These optimization functions can be extended according to the needs of an application. While single-agent shortest pathfinding can be solved in polynomial time~\cite{Dijkstra59}, \mapf with constraints on plan lengths is intractable~\cite{RatnerW86}.
%\mapf has been investigated in AI using search algorithms~\cite{Silver05,LunaB11,DresnerS08,WangB08,JansenS08,ChouhanN15,SharonSFS15,SternSFK0WLA0KB19}
%or declarative methods~\cite{YuL13,Surynek12ictai,ErdemKOS13}.

Our earlier studies~\cite{BogatarkanPE19,bogatarkan2020multi,ErdemKOS13} have addressed the challenge of flexibility for \mapf and its variants, using Answer Set Programming (ASP)~\cite{MarekT99,Niemelae99,Lifschitz02}---a logic programming paradigm based on answer sets~\cite{gelfondL91,gelfond1988stable}. In this study, we investigate the challenge of explainability for a more general variant of \mapf problem (i.e., \mmapf~\cite{bogatarkan2020multi}) applied in a robotics domain (i.e., autonomous warehouses), also utilizing ASP.

In warehouses, the robots' battery levels change as they travel around, and, in some parts of the warehouses, due to human occupancy or tight passages, the robots may need to move slowly to ensure safety. \mmapf~\cite{bogatarkan2020multi} is motivated by these realistic conditions on optimal resource use and multi-modal navigation. In \mmapf, the agents have batteries, and their battery levels change while they are moving. In the environment, there are charging stations where the agents can refill their batteries. Some parts of the environment require the agents to slow down, so agents may have different velocities depending on where they are. For example, in some tight areas, the agents may need to move slower, so it takes longer to move from one place to another. Note that these conditions regarding multi-modality make the collision constraints more complicated since agents need to traverse one edge in more than one time step. The restrictions on resource consumption and multi-modality put constraints on the routes of the agents, and thus the overall goal of completing tasks as soon as possible (i.e., minimizing the maximum plan length) or by consuming minimum energy (i.e., minimizing the total plan lengths). Both types of goals can be addressed in ASP~\cite{ErdemKOS13}.

We investigate the challenge of explainability for \mmapf problems, in particular, considering queries about the (in)feasibility and the optimality of solutions, as well as queries about the observations about these solutions.  For instance, suppose that a \mmapf solution is being executed in a warehouse. Suppose also that an engineer in this warehouse would like to check whether some modifications of this \mmapf solution would still be feasible or not.

\begin{itemize}
\item {\em Explaining infeasibility or nonoptimality.} Suppose that the modified solution is found infeasible, e.g., using the ASP methods introduced by Bogatarkan et al.~\citeyear{bogatarkan2020multi}. Then, an explanation regarding the infeasibility of the modified solution could be ``due to collisions with obstacles or other robots'', or ``due to low battery-level.'' An explanation regarding nonoptimality of the modified solution could be ``because some more time is needed to complete tasks'' or ``because some more charging is required''.
\item {\em Confirming feasibility and suggesting alternatives.} Suppose that the modified solution is found feasible. Furthermore, a better solution is computed (e.g., where the tasks are completed earlier).  Then, in addition to confirming the feasibility of the plan, it would be useful to provide the alternative solutions to the engineer.
\end{itemize}

In an alternative scenario, suppose that the engineer would like to better understand the \mmapf solution being executed in the warehouse, and asks various queries about it. For such queries, it will be useful to generate explanations using counterfactuals.

 \begin{itemize}
 \item {\em Explaining why an agent is waiting too long at a location.} Suppose that the engineer observes that the agent is waiting for a while at some location but does not move, and she wants to know why. An explanation could be that ``if the agent does not wait at that location for a while, it will collide with another robot.'' Alternatively, an explanation could be ``actually, there is no need for the agent to wait there so long, but it needs to follow a different itinerary such as ... to complete tasks on time'' or ``actually, there is no need for the agent to wait there so long, but it needs to follow a different itinerary such as ... and will be late a bit.''

\item {\em Explaining why an agent is taking a longer path.} Suppose that the engineer observes that the agent is following a path that seems rather long, and she wants to know why. An explanation could be that `if the agent does not follow that itinerary then it will collide with other robots.'' Alternatively, an explanation could be ``actually, there is no need for the agent to take a long path, but it needs to follow an alternative itinerary such as ...''.

\item {\em Explaining why an agent is charging at a distant station.} Suppose that the engineer observes that the agent is charging at a particular station that seems rather distant to its destination, and she wants to know why. An explanation could be that ``if the agent does not charge at that station, it will have to wait for other robots and thus will not be able to reach the destination on time.'' Alternatively, an explanation could be ``actually, there is no need for the agent to charge there, but it needs to follow an alternative itinerary such as ...''.

\item {\em Explaining why an agent is charging many times.} Suppose that the engineer observes that the agent is charging too many times, and she wants to know why. An explanation could be that ``the agent cannot charge less, otherwise it will not be able to reach the destination on time.'' Alternatively, an explanation could be ``actually, the agent can charge less, but it needs to follow an alternative itinerary such as ...''.

\end{itemize}

Such queries and explanations would help the engineer to better understand the strengths and weaknesses of the solution being executed, as well as the limitations of the infrastructure.

With these motivating real life scenarios, we introduce a method to generate explanations for such a variety of queries about \mmapf solutions, using the expressive formalism and efficient solvers of ASP.

%----------------------------------------------------

\section{Preliminaries}

\mmapf is a generalization of \mapf to enable multiple transportation modes and to take resource consumptions of the robots into account. We have earlier defined it as a graph problem, and introduced a flexible method to declaratively solve \mmapf and its variants using ASP~\mmapf~\cite{bogatarkan2020multi}. Let us briefly go over the definition, and highlight the parts of the \mmapf ASP program.

\smallskip\noindent{\underline{\mmapf Problem Definition}}
The input of \mmapf are
\begin{itemize}
\item
a graph $G$ characterizing the warehouse where agents move around,
\item
a set $C$ describing where charging stations are located in the warehouse,
\item
a set $S$ describing where agents can be located initially and in the end,
\item
a set $O$ denoting the parts of the environment covered by the static obstacles,
\item
a set $M$ denoting transportation modes ($\ii{slow}$ and $\ii{normal}$) of edges,
\item
a function $\ii{mode}: E \rightarrow M$ denoting the parts of the corridors where the agents should travel slowly or where they are allowed to go faster,
\item
a positive integer $n$ denoting the number of agents,
\item
a set $A$ of $n$ agents,
\item
functions $\ii{init}$ and $\ii{goal}$ describing the initial locations and the goal locations of agents,
\item
a set $B$ describing battery levels,
\item
a function $\ii{init}\_\ii{battery}: A\rightarrow B$ describing the initial battery levels of agents,
\item
a set $W_{a_i}\subseteqs V$ describing the set of waypoints for each agent $a_i$, and
\item
a positive integer $\tau$ to denote an upper bound on plan lengths.
\end{itemize}

Given these input, \mmapf asks for a plan: for each agent $a_i$, a path $P_i$ in $G$ from $\ii{init}(a_i)$ to $\ii{goal}(a_i)$, a traversal $f_i$ of this path within time $u \leqs \tau$, and a battery level function $b_i$ showing how the agent's battery level changes during the traversal. A traversal $f$ of a path $P \eqs \seq{w_{1}, w_{2}, \dots, w_{n}}$ in $G$ is understood as an onto function that maps every nonnegative integer less than or equal to $t$ to a vertex in $P$ or to $\ii{intransit}$, such that, for every $w_{l}$ and $w_{l+1}$ in $P$ and for every $x \lts t$,
\begin{itemize}
	\item if $\ii{mode}(\seq{w_l,w_{l+1}}) = \ii{normal}$ and $f(x) = w_{l}$, then $f(x{+}1) = w_{l}$ or $f(x{+}1) = w_{l+1}$.
	\item if $\ii{mode}(\seq{w_l,w_{l+1}}) = \ii{slow}$ and $f(x) = w_{l}$, then $f(x{+}1) = w_{l}$, or $ f(x{+}1) = \ii{intransit}$ and $f(x{+}2) =  w_{l+1}$.
\end{itemize}
\mmapf ensures
\begin{itemize}
	\item
about $P_i$ that all the waypoints $W_{a_i}$ are visited by the agent $a_i$ without colliding with any static obstacles $O$,
\item
%\mmapf ensures
about $f_i$ that the agents do not collide with each other while traversing their paths, and
\item
%\mmapf ensures
about $b_i$ that the agents' batteries have sufficient amount of energy (by charging at stations $C$, when needed) so that the agents can complete their plans.
\end{itemize}

For further explanations of the problem definition, we refer the reader to our earlier paper~\cite[Section~4]{bogatarkan2020multi}.

\smallskip\noindent{\underline{Solving \mmapf using ASP}}
Bogatarkan et al.~\citeyear{bogatarkan2020multi} solve \mmapf using ASP by (i) representing it as a program in ASP-Core-2 language~\cite{aspcore2}, (ii) using the ASP solver~\clingo) to find the answer sets for the program, and (iii) extracting the solutions from the answer sets, if there is an answer set.

According to the representation of \mmapf by Bogatarkan et al.~\citeyear{bogatarkan2020multi}, first plans of agents are generated recursively. Every agent {\small\tt A} starts his plan at time step {\small\tt 0} at his initial location {\small\tt X}. It can either wait at its current location {\small\tt X} (if {\small\tt X} denotes a vertex but not {\small\tt intransit}) until the next time step {\small\tt T+1}, or move to the adjacent vertex {\small\tt Y} via a normal edge or a slow edge. For instance, the traversal of an edge with slow mode is described as follows:
{\small\begin{verbatim}
{plan(A,T+1,intransit)}1 :-
  plan(A,T,X), edge(X,Y), mode(X,Y,`s'), time(T), T<t-1.
1{plan(A,T+2,Y): edge(X,Y), mode(X,Y,`s')}1 :-
  plan(A,T+1,intransit), plan(A,T,X), time(T), T<t-1.
\end{verbatim}}
The uniqueness and existence of paths are ensured by constraints.

Similarly, the battery level of an agent is defined recursively. At each step {\small\tt T}, if the agent is not at a charging station, its battery level reduces by 1. If the agent is at a charging location, its battery level may quickly get to the maximum level $b$ or the agent can move forward without charging its battery. This behaviour of an agent at a charging station is described as follows:
{\small\begin{verbatim}
1{batteryLevel(A,T+1,b); batteryLevel(A,T+1,B1-1)}1 :-
  plan(A,T,X), batteryLevel(A,T,B1), charging(X), agent(A),
  time(T), T<L, planLength(A,L).
\end{verbatim}}
Constraints are added to ensure that a minimum level of battery level.

After that, \mmapf constraints are formulated. For instance, the collision constraint ``No two agents are at the same place at the same time, except when they are both in transit.'' are described as follows:
{\small\begin{verbatim}
:- plan(A1,T,X), plan(A2,T,X), agent(A1;A2), A1<A2, X!=intransit.
\end{verbatim}}

For details of the ASP formulation, we refer the reader to our earlier paper~\cite[Section~5]{bogatarkan2020multi}.

%----------------------------------------------------

\section{Explanation Generation for \mmapf using ASP}

Our ASP-based method consists of two parts: the main algorithm (implemented using Python), and the main ASP program $\Pi$ for \mmapf (represented in ASP-Core-2 language, as described by Bogatarkan et al.~\citeyear{bogatarkan2020multi}).

\begin{algorithm}
\SetKwInOut{Input}{Input}\SetKwInOut{Output}{Output}
\Input{a \mmapf instance, a plan for this instance, and a query $q$ of type QW1--QU}
\Output{An explanation}
// Suppose that $\Pi$ denotes the \mmapf program described by Bogatarkan et al.~\citeyear{bogatarkan2020multi}, possibly augmented with some hard constraints due to previous queries \\
%\tcp{Suppose that $\Pi$ denotes the \mmapf program described in~\cite{bogatarkan2020multi}}
\If {query $q$ is of type QW1--QP5}{
	$\Pi_h\gets$ Add the relevant hard constraint for $q$ to the \mmapf program $\Pi$ \\
	\If {$\Pi_h$ has an answer set $X$}{
        Display an explanation, presenting an alternative (better/worse) solution
%		\If{$X$ has the same optimization values as the given plan}{
%			Display \emph{``There is another solution: ...''}
%		}
%		\ElseIf{$X$ has better optimization values than the given plan}{
%			Display \emph{``There is another solution with better optimization values: ...''}	
%		}
%		\Else%{$X$ has worse optimization values than the given plan}{
%		{
%			Display \emph{``There is another solution but with worse optimization values: ...''}
%		}
	}
	\Else{
		$\Pi_w\gets$ Replace the \mmapf constraints in $\Pi_h$ relevant for $q$, with the corresponding rules and weak constraints \\
		$Y\gets$ Compute an answer set for $\Pi_w$ \\
		Display a counterfactual-guided explanation, based on which constraints are violated
	}
}
\Else{ // query $q$ is of type QU \\
	$\Pi_w\gets$ Replace the \mmapf constraints in $\Pi$ with the corresponding rules and weak constraints \\
	$Y\gets$ Compute an answer set for $\Pi_w$ \\
	Display a counterfactual-guided explanation, based on which constraints are violated
}

\caption{The main algorithm for generating explanations for \mmapf problems.}
\label{algo1}
\end{algorithm}

The inputs of the main algorithm (Algorithm~\ref{algo1}) are a \mmapf instance, a plan for this instance, and a query. Currently, our algorithm supports 14 types of queries about the given plan. Queries QW1--QW4 are about waiting, QC1--QC4 are about charging, QP1--QP5 are about traversals, and QU is about the nonexistence of a solution.

\begin{enumerate}
    \item[QW1] Why does Agent {\small\tt a} wait at location {\small\tt x} (at any time)?
    \item[QW2] Why does Agent {\small\tt a} wait at location {\small\tt x} at time {\small\tt s}?
    \item[QW3] Why does Agent {\small\tt a} wait at location {\small\tt x} at time {\small\tt s} for {\small\tt n} steps?
    \item[QW4] Why does not Agent {\small\tt a} wait at location {\small\tt x} at time {\small\tt s} for less than {\small\tt n} steps?

    \item[QC1] Why does Agent {\small\tt a} charge at location {\small\tt x} (at any time)?
    \item[QC2] Why does Agent {\small\tt a} charge at time {\small\tt s}?
    \item[QC3] Why does Agent {\small\tt a} charge at location {\small\tt x} at time {\small\tt s}?
    \item[QC4] Why does not Agent {\small\tt a} charge less than {\small\tt m} times?

    \item[QP1] Why does not Agent {\small\tt a} have a plan whose length is less than {\small\tt l}?
    \item[QP2] Why does Agent {\small\tt a} visit location {\small\tt x} (at any time)?
    \item[QP3] Why does Agent {\small\tt a} visit location {\small\tt x} at time {\small\tt s}?
    \item[QP4] Why does Agent {\small\tt a} move from location {\small\tt x} to location {\small\tt y} (at any time)?
    \item[QP5] Why does Agent {\small\tt a} move from location {\small\tt x} to location {\small\tt y} at time {\small\tt s}?

    \item[QU] Why does not the instance have a solution?
\end{enumerate}

\subsection{Explanation Generation for Queries QW1--QP5 about Solutions}

Queries QW1--QP5 are associated by the following hard constraints:

\begin{enumerate}
    \item[QW1] {\small \verb|:- plan(a,T,x), plan(a,T+1,x), T<t.|}
    \item[QW2] {\small \verb|:- plan(a,s,x), plan(a,s+1,x).|}
    \item[QW3] {\small \verb|:- plan(a,s,x),..,plan(a,s+n,x).|}
    \item[QW4] {\small \verb|:- C = #count{T: plan(a,T,x), plan(a,T+1,x), T<s+n, T>=s}, C >= n.|}
    \item[QC1] {\small \verb|:- batteryLevel(a, T+1, b), plan(a, T, x), charging(x).|}
    \item[QC2] {\small \verb|:- batteryLevel(a,s+1,b), plan(a,s,X), charging(X).|}
    \item[QC3] {\small \verb|:- batteryLevel(a, s+1, b), plan(a, s, x), charging(x).|}
    \item[QC4] {\small \verb|:- C = #count{T: batteryLevel(a,T,b), T>0}, C >= m.|}

    \item[QP1] {\small \verb|:- planLength(a,L), L>=l.|}
    \item[QP2] {\small \verb|:- plan(a,T,x).|}
    \item[QP3] {\small \verb|:- plan(a,s,x).|}
    \item[QP4] {\small \verb|:- plan(a,T,x), plan(a,T+1,y), edge(x,y), mode(x,y,`n'), T<t-1.|}
    \item[]    {\small \verb|:- plan(a,T,x), plan(a,T+1,intransit), plan(a,T+2,y),|}
    \item[]    {\small \verb|   edge(x,y), mode(x,y,`s'), T<t.|}
    \item[QP5] {\small \verb|:- plan(a,s,x), plan(a,s+1,y), edge(x,y), mode(x,y,`n').|}
    \item[]    {\small \verb|:- plan(a,s,x), plan(a,s+1,intransit), plan(a,s+2,y),|}
    \item[]    {\small \verb|   edge(x,y), mode(x,y,`s').|}
\end{enumerate}

If the given query is of type QW1--QP5, then the algorithm first checks whether the ASP program $\Pi_h$, obtained from $\Pi$ by adding the relevant hard constraint, has an answer set or not. If the augmented program has an answer set, then alternative plans are extracted from the answer sets and presented to the user with some recommendations. Here as some sample explanations for some queries:

\begin{enumerate}
    \item[QW1] Actually, Agent {\small\tt a} does not have to wait at location {\small\tt x}. Here is an alternative plan: ...
    \item[QW3] Actually, Agent {\small\tt a} does not have to wait at location {\small\tt x} at time {\small\tt s} for {\small\tt n} steps. Here is an alternative plan: ...
    \item[QC2] Actually, Agent {\small\tt a} does not have to charge at time step {\small\tt s}. Here is an alternative plan: ...
    \item[QC4] Actually, Agent {\small\tt a} can charge less than {\small\tt m} times. Here is an alternative plan: ...
    \item[QP1] Actually, Agent {\small\tt a} can follow a shorter path whose length is smaller than {\small\tt l}. Here is an alternative plan: ...
    \item[QP5] Actually, Agent {\small\tt a} does not have to move from location {\small\tt x} to location {\small\tt y} at time {\small\tt s}. Here is an alternative plan: ...
\end{enumerate}

\noindent If the given plans are optimal then the explanations can involve further information: ``Here is an alternative plan that is shorter: ....''

If the augmented program does not have an answer set, then the given plan is not feasible or optimal. Then the algorithm finds an explanation for why not, i.e., a query of type QU.

\subsection{Explanation Generation for Query QU about the Nonexistence of Solutions}

If a given plan is found infeasible or nonoptimal, our algorithm tries to identify which constraints relevant to the given question are violated. For that reason, we obtain a new ASP program $\Pi_w$ from the \mmapf program $\Pi_h$ by replacing each relevant \mmapf constraint by a set of rules and a weak constraint as follows.

We replace the collision constraint (i.e., no two agents are at the same place at the same time, except when they are both in transit)
{\small\begin{verbatim}
:- plan(A1,T,X), plan(A2,T,X), agent(A1;A2), A1<A2, X!=intransit.
\end{verbatim}}
\noindent by the following rules that describe the conditions violating this collision constraint:
{\small\begin{verbatim}
violate_collision(A1,A2,T,X) :-
  plan(A1,T,X), plan(A2,T,X), agent(A1;A2), A1<A2, X!=intransit.
\end{verbatim}}
\noindent and the following weak constraints:
{\small\begin{verbatim}
:~ violate_collision(A1,A2,T,X). [1@7, A1,A2,T,X,vc]
\end{verbatim}}

We replace the swapping constraints (i.e., swapping is not allowed along a normal edge or a slow edge)
{\small\begin{verbatim}
:- plan(A1,T,X), plan(A1,T+1,Y), plan(A2,T,Y), A1<A2,
   plan(A2,T+1,X), agent(A1;A2), mode(X,Y,`n'), T<t.
:- slow(A1,T,X,Y), slow(A2,T-1,Y,X), T>0, T<t-1, A1!=A2.
:- slow(A1,T,X,Y), slow(A2,T,Y,X), T<t-1, A1<A2.
\end{verbatim}}
\noindent by the following rules that describe the conditions violating these swapping constraints:
{\small\begin{verbatim}
violate_swap(A1,A2,T,X,Y) :-
   plan(A1,T,X), plan(A1,T+1,Y), plan(A2,T,Y),
   A1<A2, plan(A2,T+1,X), agent(A1;A2), mode(X,Y,`n'), T<t.
violate_slow_collision1(A1,A2,T,X,Y):-
   slow(A1,T,X,Y), slow(A2,T-1,Y,X), T>0, T<t-1, A1!=A2.
violate_slow_collision2(A1,A2,T,X,Y):-
   slow(A1,T,X,Y), slow(A2,T,Y,X), T<t-1, A1<A2.
\end{verbatim}}
\noindent and the following weak constraints:
{\small\begin{verbatim}
:~ violate_swap(A1,A2,T,X,Y). [1@7, A1,A2,T,X,Y,vs]
:~ violate_slow_collision1(A1,A2,T,X,Y). [1@7, A1,A2,T,X,Y,vsc1]
:~ violate_slow_collision2(A1,A2,T,X,Y). [1@7, A1,A2,T,X,Y,vsc2]
\end{verbatim}}

Similarly, we replace the goal constraint (i.e., the agent should reach its destination), the waypoint constraint (i.e., the agent should visit the waypoints in its way to its destination), the obstacle collision constraint (i.e., no agent collides with an obstacle), and the battery constraint (i.e., the agent should have a positive battery level) by the following rules:
{\small\begin{verbatim}
violate_goal(A,X) :- goal(A,X), not visit(A,X).
violate_waypoint(A,X) :- waypoint(A,X), not visit(A,X).
violate_obstacle(A,T,X) :- plan(A,T,X), obstacle(X), agent(A), time(T).
violate_min_battery(A,T,X):- batteryLevel(A,T,0), plan(A,T,X),
  planLength(A,L), T<L.
\end{verbatim}}
\noindent and the following weak constraints:
{\small\begin{verbatim}
:~ violate_goal(A,X). [1@7, A,X,vg]
:~ violate_waypoint(A,X). [1@7, A,X,vw]
:~ violate_obstacle(A,T,X). [1@7, A,T,X,vo]
:~ violate_min_battery(A,T,X). [1@7, A,T,vb]
\end{verbatim}}

The idea is to identify from an answer set for $\Pi_w$, which constraints are violated, and then present an explanation to the user accordingly.

\subsection{Discussions}
We briefly discuss some useful extensions and capabilities of our algorithm.

\smallskip\noindent{\underline{Relevancy of constraints to questions}}
Instead of considering all constraints, we identify the constraints relevant to the given question to generate more meaningful explanations. For QW type queries, the constraints about collisions, obstacle, and waypoints are relevant. For QC type queries, the constraints about battery level, goal, obstacle, and waypoints are relevant. For QP and QU queries, all constraints (about battery level, collisions, goal, obstacle, waypoints) are relevant.

\smallskip\noindent{\underline{Weights and priorities}} In the ASP formulation presented above, the priority of the weak constraints is specified as a number larger than the priority of weak constraints used for \mmapf optimizations (i.e., minimizing the total number of times of charging, or minimizing the total plan length) because \mmapf constraints are more important than optimizations.  Meanwhile, all \mmapf constraints are considered equally important, with the same priority and the same weight. In real-world applications, the users can change these priorities.

\smallskip\noindent{\underline{$\Pi_h$ vs. $\Pi_w$}} Our algorithm first tries to generate an explanation for the observations of the user over the given plan, by means of queries QW1--QP5, using the program $\Pi_h$. If the given plan is not feasible or optimal, it generates further explanations by  utilizing weighted weak constraints, i.e., using the program $\Pi_w$. The scrupulous reader might notice that we could have used $\Pi_w$ from the very beginning, to answer queries QW1--QP5. However, it would not be computationally efficient. For instance, for a query of type QP1 over an instance with 4 agents on a grid of size $22\times 22$, %(Figure~\ref{fig:fig4}),
finding an answer using the program $\Pi_h$ takes $31$ seconds, while it takes $2249$ seconds using the program $\Pi_w$.

\smallskip\noindent{\underline{Additional explanations for QW and QC queries}}
If the augmented program $\Pi_h$ does not have an answer set, then the given plan is not feasible or optimal. At this point, for QW and QC queries about waiting and charging, the algorithm can generate further explanations by answering the following question: ``What would happen if the agent did not wait/charge in the given plan?'' For that, the algorithm revises the plan to exclude waiting/charging, obtains a program $\Pi_c$ from program $\Pi_w$ by adding the revised plan as a hard constraint, and checks the answer sets for $\Pi_c$. Scenario 2 in the next section provides a good example.

\smallskip\noindent{\underline{Possibility of infrastructure change}}
For queries QU, our algorithm can generate further explanations based on the possibility of infrastructure change (e.g., removing some shelves) in the warehouse. For that purpose, the algorithm obtains two programs $\Pi_{w1}$ and $\Pi_{w2}$ from $\Pi_h$. If an infrastructure change is not possible, $\Pi_{w1}$ does not include the weak constraint for obstacles; instead, it includes the hard constraints for obstacles and other relevant weak constraints. If an infrastructure change in the warehouse is possible, $\Pi_{w2}$ includes the weak constraint for obstacles only. The algorithm computes answer sets for each program and generates more comprehensive explanations.  Scenario 6 in the next section provides a good example.
%
%\begin{figure}[t]
%  \centering
%  \includegraphics[scale=0.3]{largeWithoutAgents.pdf}
%  \caption{$22x22$ grid}
%  \label{fig:fig4}
%\end{figure}

%----------------------------------------------------

\section{Examples}

\begin{figure}[t]
\begin{tabular}{cccc}
\centering
        \resizebox{0.22\textwidth}{!}{\begin{oldtabular}{c}
        \includegraphics[scale = 0.3]{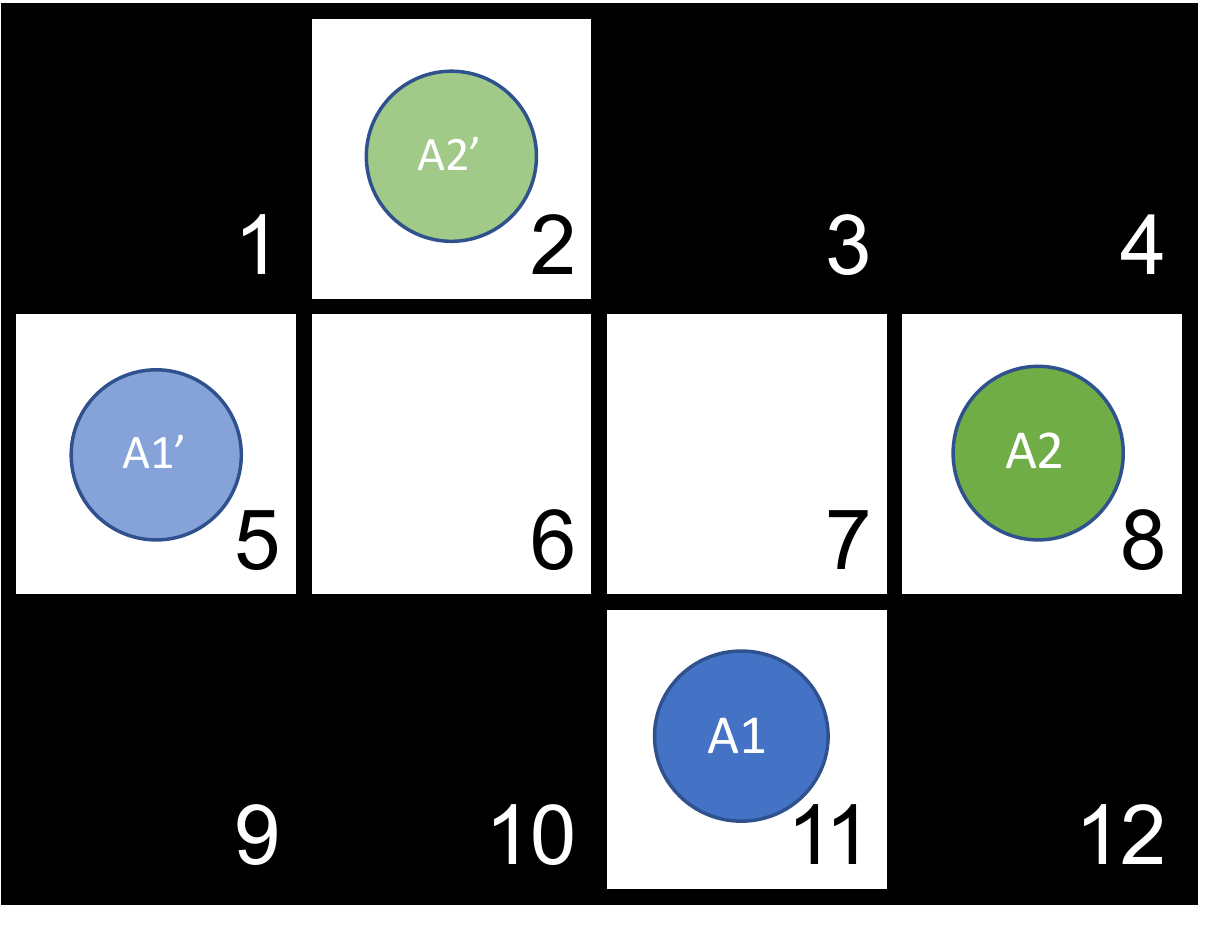}
        \end{oldtabular}}
        &
        \resizebox{0.2\textwidth}{!}{\begin{oldtabular}{ccc}
        \hline\hline
        Time & A1 Location & A2 Location \\ \hline%\begin{tabular}[c]{@{}c@{}}A1 \\ Location\end{tabular} & \begin{tabular}[c]{@{}c@{}}A2 \\ Location\end{tabular} \\ \hline
        0 & 11 & 8 \\
        1 & 7 & 8 \\
        2 & 6 & 7 \\
        3 & 5 & 6 \\
        4 & - & 2 \\  \hline\hline
        \end{oldtabular}}
        &
        \resizebox{0.2\textwidth}{!}{\begin{oldtabular}{ccc}
        \hline\hline
        Time & A1 Location & A2 Location \\ \hline % \begin{tabular}[c]{@{}c@{}}A1 \\ Location\end{tabular} & \begin{tabular}[c]{@{}c@{}}A2 \\ Location\end{tabular} \\ \hline
        0 & 11 & 8 \\
        1 & 11 & 7 \\
        2 & 7 & 6 \\
        3 & 6 & 2 \\
        4 & 5 & - \\ \hline\hline
        \end{oldtabular}}
        &
        \resizebox{0.25\textwidth}{!}{\begin{oldtabular}{c}
        \includegraphics[scale=0.3]{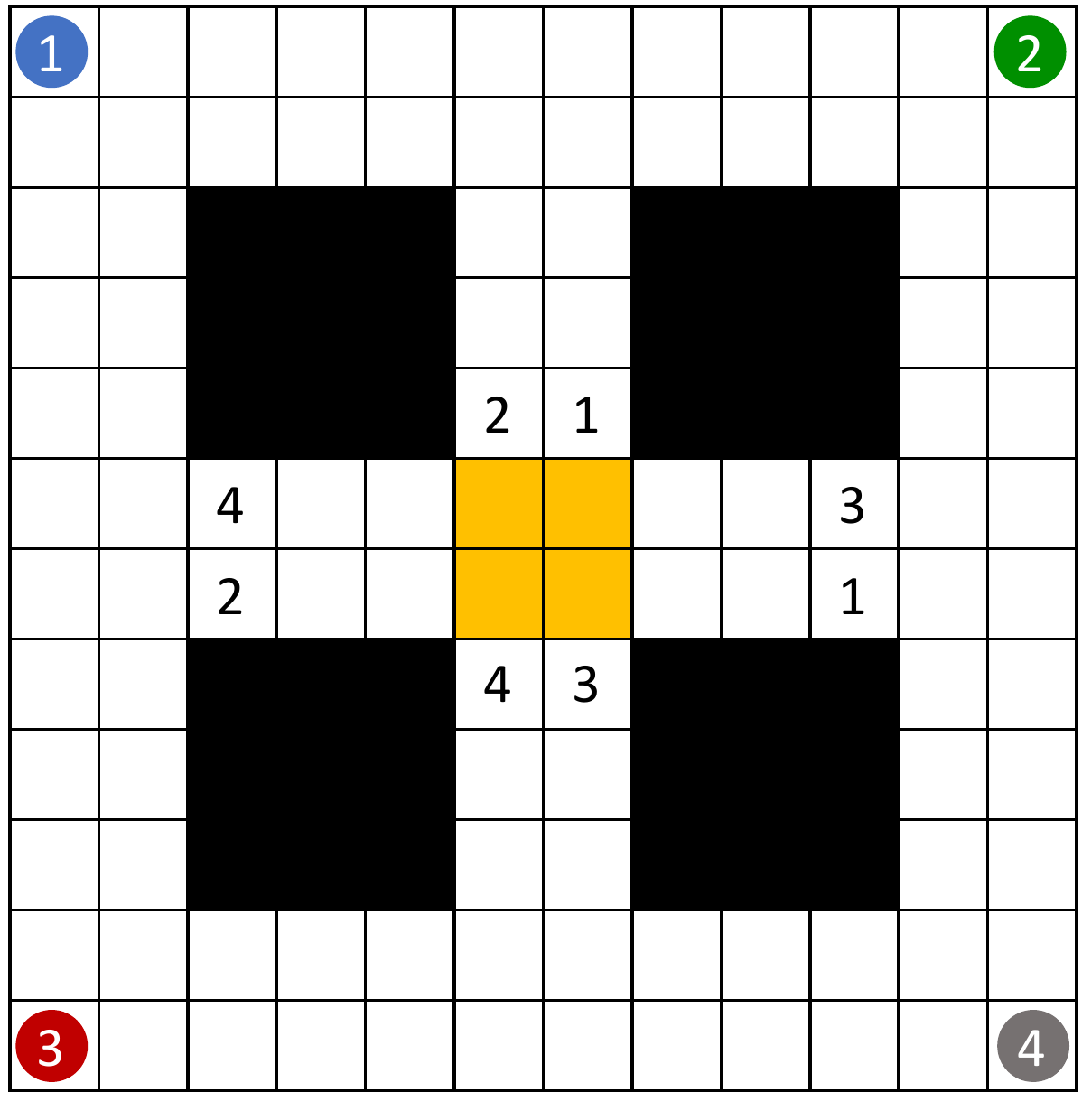}
        \end{oldtabular}}
        \\
        (a) & (b) & (c) & (d) \\
        \end{tabular}
            \vspace{-\baselineskip}
        \caption{\footnotesize(a) Scenarios~1 and~2: A1 and A2 denote the initial positions of Robots~1 and~2; A1$'$ and A2$'$ denote goal locations. Cell~7 is a waypoint for both robots, (b) and (c) are two optimal plans for this instance. (d) Scenario~3: Robots~1--4 are initially located at different corners of a $12x12$ grid; each robot aims to reach the diagonally-opposite corner. Other cells labeled by a number $i$ are waypoints for Robot~$i$. Yellow cells are charging stations and black cells are obstacles.}
        \label{fig:fig1}
            \vspace{-\baselineskip}
    \end{figure}

\smallskip\noindent{\underline{Scenario 1}}
Consider the \mmapf instance shown in Figure~\ref{fig:fig1}(a) in a small warehouse, where Robot~1 is initially located at Cell~11 and aims to reach Cell~5, and Robot~2 is initially located at Cell~8 and aims to reach Cell~2. Cell~7 is a waypoint for both robots, and the upper bound on makespan of a plan is 4. For simplicity, suppose that the batteries of the robots are fully charged initially, and sufficient for execution of any given plan, and that all edges have the normal mode of transportation.

About Plan~1 described in Figure~\ref{fig:fig1}(b), suppose that an engineer asks the following query of type QW1:
\begin{quote}
``Why does Robot~2 wait at Cell~8 (at any time)?''
\end{quote}

Our algorithm first obtains $\Pi^1_h$ by adding the following constraint to the \mmapf program $\Pi$:
{\small\begin{verbatim}
:- plan(2,T,8), plan(2,T+1,8), T<t.
\end{verbatim}}
\noindent checks whether there is an optimal plan where Robot~2 does not have to wait initially (i.e., $\Pi_h$ has an answer set).

Once an alternative solution, Plan~2 described in Figure~\ref{fig:fig1}(c), is found, our algorithm presents the following explanation: \begin{quote}
``Actually, Robot~2 does not have to wait at Cell~8 from time step 0 to 2. Here is an alternative optimal plan: Plan~2...''
\end{quote}

\smallskip\noindent{\underline{Scenario 2}} Continuing Scenario~1, suppose that the engineer then asks the following query of type QW1:
\begin{quote}
``Why does Robot~1 wait at Cell~11 (at any time)?''
\end{quote}

Our algorithm first obtains $\Pi^2_h$ by adding the following constraint to $\Pi^1_h$:
{\small\begin{verbatim}
:- plan(1,T,11), plan(1,T+1,11), T<t.
\end{verbatim}}
\noindent and checks whether there is an optimal plan where Robot~1 does not have to wait initially (i.e., $\Pi^2_h$ has an answer set). However, $\Pi^2_h$ does not have an answer set: there are no other solutions.

Then, our algorithm tries to generate explanations by answering two questions: ``What would happen if Robot~1 does not wait at Cell~11 in the current plan?'' ``What will happen if Robot~1 does not wait at Cell~11?'' To answer the first question, our algorithm obtains a program $\Pi^2_w$ from $\Pi^2_h$, by replacing the collision constraints with the relevant rules and weak constraints as described above. It revises the current plan so that Robot~1 does not wait at Cell~11 (and the plans of other agents do not change), and obtains a program $\Pi^2_c$ by adding the revised plan as a hard constraint to $\Pi^2_w$. After that, our algorithm computes an answer set for $\Pi^2_c$, identifies the atoms {\small\tt violate\_collision(1,2,1,7)} and {\small\tt violate\_collision(1,2,2,6)} in it, generates the following explanation for the first question:  \begin{quote}
``Robot~1 has to wait at Cell~11 in the current plan; otherwise, Robot~1 and Robot~2 would collide with each other at Cell~7 at time step~1 and at Cell~6 at time step~2.''
\end{quote}
%After that, our algorithm computes an answer set for $\Pi^2_c$  and identifies the following atoms in the answer set:
%{\small\tt violate\_collision(1,2,1,7)}, {\small\tt violate\_collision(1,2,2,6)}.
%This answer set describes a plan (with a makespan of at most 3) for Robots~1 and~2, according to which
%they collide with each other at Cell~7 at time step~1 and at Cell~6 at time step~2.
%Then, our algorithm presents the following explanation to the engineer:
%\begin{quote}
%``Robot~1 has to wait at Cell~11; otherwise, Robot~1 and Robot~2 will collide with each other at Cell~7 at time step~1 and at Cell~6 at time step~2 when they follow the given plans.''
%\end{quote}

To answer the second question, our algorithm finds an answer set for $\Pi^2_w$. This answer set contains a different plan with a makespan of 4 and the atom {\small\tt violate\_collision(1,2,1,7)}, and thus our algorithm generates the following explanation for the second question:
\begin{quote}
``Robot~1 has to wait at Cell~11; otherwise, Robot~1 and Robot~2 will collide with each other at Cell~7 with another plan.''
\end{quote}

%Finally, our algorithm presents the following explanation to the engineer:
%\begin{quote}
%	``Robot~1 has to wait at Cell~11; otherwise, Robot~1 and Robot~2 will collide with each other at Cell~7 at time step~1 and at Cell~6 at time step~2 when they follow the given plans or Robot~1 and Robot~2 will collide with each other at Cell~7 with another plan.''
%\end{quote}

%Then, it computes an answer set for $\Pi^2_w$, that includes the atom
%{\small\begin{verbatim}
%violate_collision(1,2,1,7).
%\end{verbatim}}
%This answer set describes a plan (with a makespan of at most 4) for Robots~1 and~2, according to which the robots collide with each other at Cell~7 at time step~1.
%
%Then, our algorithm presents the following explanation to the engineer:
%\begin{quote}
%``Robot~1 has to wait at Cell~11 from timestep 0 to 2; otherwise, Robots 1 and 2 will collide with each other at Cell~7 at time step~1.''
%\end{quote}

\smallskip\noindent{\underline{Scenario 3}}
Consider the \mmapf instance shown in Figure~\ref{fig:fig1}(d).
Robots~1--4 are initially located at different corners of the warehouse. Each robot aims to reach the diagonally-opposite corner. Other cells labeled by a number $i$ are waypoints for Robot~$i$. Yellow cells are charging stations and black cells are obstacles.

Suppose that a plan is given to an engineer as an optimal solution, where each Robot executes a plan of length 22, and she asks a query of type~QP1:
\begin{quote}
``Why does not Robot~1 follow a shorter plan whose length is smaller than 22?''
\end{quote}

%\begin{figure}[t]
%    \centering
%    \includegraphics[scale=0.51]{smallWithAgents.pdf}
%    \caption{Scenario~3: Robots~1--4 are initially located at different corners of a $12x12$ grid; each robot aims to reach the diagonally-opposite corner. Other cells labeled by a number $i$ are waypoints for Robot~$i$. Yellow cells are charging stations and black cells are obstacles.}
%    \label{fig:fig2}
%\end{figure}

Our algorithm first obtains the program $\Pi_h$ from the \mmapf program $\Pi$ by adding the constraint
{\small\begin{verbatim}
:- planLength(1,L), L>=22.
\end{verbatim}}
\noindent and checks whether there is another optimal solution where Robot~1 follows a shorter plan.

Since the program $\Pi_h$ does not have an answer set, our algorithm tries to identify which \mmapf constraint is violated. For that, it obtains the program $\Pi_w$ from $\Pi_h$ by replacing all \mmapf constraints with relevant rules and weak constraints, as described in the previous section.

According to an answer set computed for $\Pi_w$, which contains
%the atoms
{\small\tt violate\_waypoint(1,55)}, {\small\tt violate\_waypoint(1,82)} and {\small\tt violate\_goal(1,144)},
the algorithm generates the following explanation:
\begin{quote}
``Robot~1 cannot follow a shorter plan; otherwise, it would not be possible to visit all the waypoints and to reach its goal.''
\end{quote}

%?????????????????????????
%
%
%We cannot remove goal from the relevant constraints of plan length query
%otherwise we cannot find a plan even with the violated constraints since a shorter plan is never possible because of the size of the grid. We can change the sentence in the explanation by only mentioning goal in the explanation:
%
%According to an answer set computed for $\Pi_w$, that contains the atoms
%{\small\begin{verbatim}
%violate_waypoint(1,55) violate_waypoint(1,82) violate_goal(1,144)
%\end{verbatim}}
%the algorithm generates the following explanation:
%\begin{quote}
%``Robot~1 cannot follow a shorter plan; otherwise, it would not be possible to reach its goal.''
%\end{quote}

\begin{figure}[t]
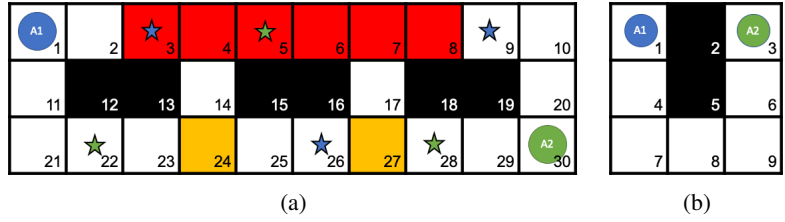

\begin{tabular}{cc}
\centering
        %\resizebox{0.2\textwidth}{!}{\begin{oldtabular}{c}
        \includegraphics[scale=0.25]{qc4_instance.pdf}
        %\end{oldtabular}}
        &
        %\resizebox{0.2\textwidth}{!}{\begin{oldtabular}{ccc}
        \includegraphics[scale=0.25]{qu_instance.pdf} \\
        (a) & (b)
        %\end{oldtabular}}
\end{tabular}
    \vspace{-\baselineskip}
    \caption{\footnotesize(a) Instance M1. Robots~1 and~2 are initially located at two corners of the grid and each robot aims to reach the diagonally-opposite corner. The stars show the waypoints of the robot with the same color. Yellow cells are charging stations, red cells are the slow zone and black cells are obstacles. (b) Robots~1 and~2 are initially located at the given positions on the grid; the goal of each robot is the initial location of the other. Black cells are obstacles.}
    \label{fig:fig_qc4}
%    \vspace{-\baselineskip}
\end{figure}

\smallskip\noindent{\underline{Scenario 4}}
Let us consider the \mmapf instance shown in Figure~\ref{fig:fig_qc4}(a), Robots~1--2 are initially located at $1$ and $30$ an their goal locations are $30$ and $1$, respectively. Yellow cells are charging stations, red cells are the slow zone and black cells are obstacles. The stars show the waypoints of the agent of the same color. Maximum battery level is $10$ and Robot~1 is fully charged initially but initial battery level of Robot~2 is $8$.

Suppose that a plan is given to an engineer as an optimal solution, where each Robot is charged $2$ times in the plan. The engineer wants to charge less, so she asks the following query of type~QC4:
\begin{quote}
`` Why does not Robot~2 charge less than $2$ times?''
\end{quote}

To answer this query, our algorithm adds the following constraint to the \mmapf program $\Pi$ and obtains $\Pi_h$:
{\small\begin{verbatim}
:- C = #count{T: batteryLevel(2,T,b), T>0} , C >= 2.
\end{verbatim}}
Then, it tries to find a plan with less number of charging (i.e., an answer set for $\Pi_h$). There is no answer set of $\Pi_h$. The algorithm obtains the program $\Pi_w$ from $\Pi_h$ by replacing the relevant hard constraints with the weak constraints, and then obtains the program $\Pi_c$ from $\Pi_w$ by adding the current traversals of the agents as hard constraints. The answer set for $\Pi_c$ contains the atom
{\small\tt violate\_min\_battery(2,8,intransit)}.

Then, our algorithm further tries to find an answer set for $\Pi_w$ to find out which constraints would be violated regardless of the current plan. The answer set contains a different plan and the atom {\small\tt violate\_waypoint(2,5)}.

In the end, the algorithm generates the following explanation:
\begin{quote}
 ``Robot~2 cannot charge less than $2$ times; otherwise, its battery will run out at time step~8 if it uses the current plan or it will not be able to visit its waypoint at Cell~5 with another plan.''
\end{quote}

\smallskip\noindent{\underline{Scenario 5}}
Again, let's consider the scenario in Figure~\ref{fig:fig_qc4}(a). Robot~1 has the plan $$P_1=\seq{1,2,3,intransit,4,14,24,25,26,27,17,7,intransit,8,9,10,20,30}.$$
The engineer wants to know if there is a plan without visiting the edge $\seq{4,14}$
and asks the following query of type~QP4:
\begin{quote}
``Why does Robot~1 move from Cell~4 to Cell~14 (at any time)?''
\end{quote}

The algorithm adds the following constraint to $\Pi$ and obtains $\Pi_h$:
{\small\begin{verbatim}
:- plan(1,T,4), plan(1,T+1,14), edge(4,14), mode(4,14,`n'), T<t.
\end{verbatim}}

There exists an answer set for $\Pi_h$, with a longer plan for Robot~1. Therefore, the algorithm generates the following explanation:
\begin{quote}
 ``Actually, Robot~1 does not have to move from Cell~4 to Cell~14. Here is an alternative plan which is longer: ...
\end{quote}

\smallskip\noindent{\underline{Scenario 6}}
Now consider the instance in Figure~\ref{fig:fig_qc4}(b). Robots~1--2 are initially located at $1$ and $3$; their goals are at $3$ and $1$, respectively. Black cells denote obstacles. For simplicity, we assume that their batteries are initially fully charged and enough to traverse all of their paths, all edges have normal mode of transportation, and the only waypoint of each robot is located at its initial position.

%\begin{figure}[ht]
%    \centering
%    \includegraphics[scale=0.51]{qu_instance.pdf}
%    \caption{Robots~1--2 are initially located at the given positions on the grid; the goal of each robot is the initial location of the other. Black cells are obstacles.}
%    \label{fig:fig_qu}
%\end{figure}

There is no solution for this \mmapf instance. Suppose that an engineer wants to find out the reason and asks query QU:
\begin{quote}
``Why does not the instance have a solution?''
\end{quote}

The algorithm obtains two programs $\Pi_{w1}$ and $\Pi_{w2}$ from $\Pi_h$, as explained in the previous section, and tries to find an answer set for each program. The answer set for $\Pi_{w1}$ contains {\small\tt violate\_collision(1,2,3,8)}, and the algorithm generates the following explanation:
\begin{quote}
``There is no solution because Robot~1 and Robot~2 collide at Cell~8 at time step~3.''
\end{quote}

\noindent The answer set for $\Pi_{w2}$ contains {\small\tt violate\_obstacle(2,1,2)}, and the algorithm further generates the following explanation:
\begin{quote}
``There is no solution because Robot~2 collides with the obstacle at Cell~2 at time step~1; this suggests removing this obstacle.''
\end{quote}

%----------------------------------------------------
%\vspace{-2ex}

\section{Experimental Evaluations}

\begin{figure}[t]
	\centering
	\includegraphics[scale=0.25]{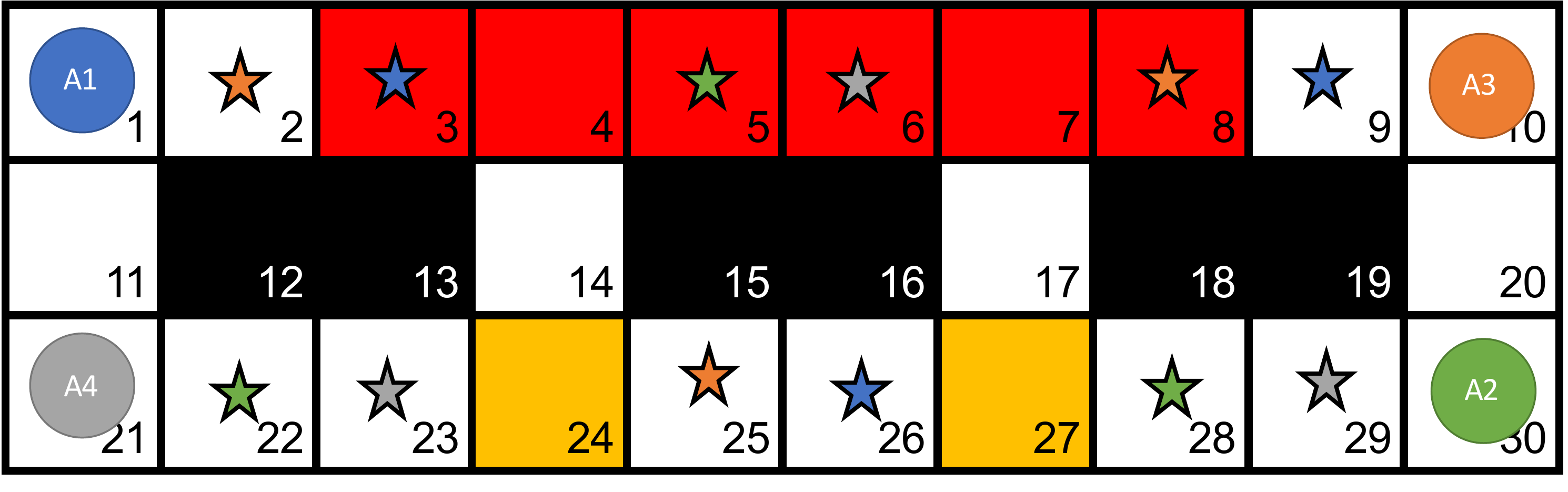}
    \vspace{-\baselineskip}
	\caption{\footnotesize Instance M2 with 4 agents initially located at corners and their goals are located at the diagonally-opposite corner.}
	\label{fig:fig_m2}
    %\vspace{-\baselineskip}
\end{figure}

\begin{figure}[t]
	\centering
	\includegraphics[width=\textwidth]{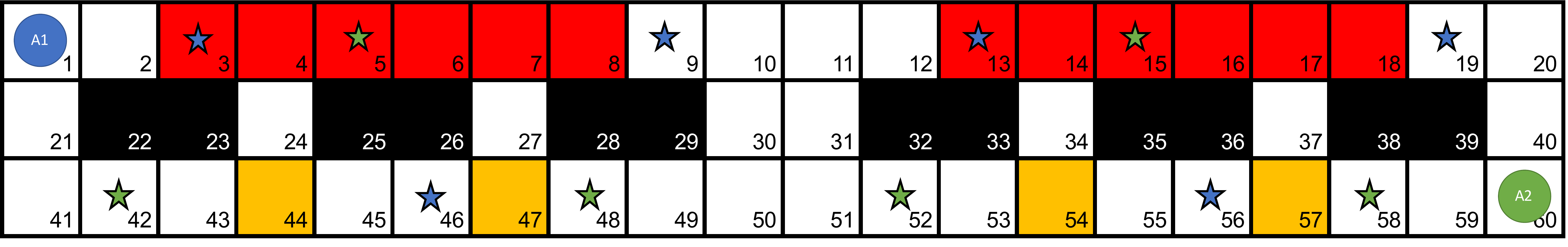}
    \vspace{-\baselineskip}
	\caption{\footnotesize Instance M3 with 2 agents an a larger grid. Agents are located at two corners of the grid and their goals are at the diagonally-opposite corner.}
	\label{fig:fig_m3}
%    \vspace{-\baselineskip}
\end{figure}

%\begin{figure}[t]
%\begin{tabular}{cc}
%\centering
%        %\resizebox{0.2\textwidth}{!}{\begin{oldtabular}{c}
%        \includegraphics[scale=0.25]{qc4_instance.pdf}
%        %\end{oldtabular}}
%        &
%        %\resizebox{0.2\textwidth}{!}{\begin{oldtabular}{ccc}
%        \includegraphics[scale=0.25]{qu_instance.pdf} \\
%        (a) & (b)
%        %\end{oldtabular}}
%\end{tabular}
%    \vspace{-\baselineskip}
%    \caption{\footnotesize(a) Robots~1--2 are initially located at two corners of the grid and each robot aims to reach the diagonally-opposite corner. The stars show the waypoints of the robot with the same color. Yellow cells are charging stations, red cells are the slow zone and black cells are obstacles. (b) Robots~1--2 are initially located at the given positions on the grid; the goal of each robot is the initial location of the other. Black cells are obstacles.}
%    \label{fig:fig_qc4}
%    \vspace{-\baselineskip}
%\end{figure}

We have evaluated our ASP-based method for generating explanations, considering all 13 types of queries (except query QU) over two \mmapf instances with multi-modality, resource, waypoint, and plan length constraints within a tight space to move around. The first instance M1 is shown in Figure~\ref{fig:fig_qc4}(a), with two agents and where the upper bound on makespan is 18. The second \mmapf instance M2 considers the same environment as in the first instance, with two more agents initially placed in the empty corners (Cells 10 and 21) with the goal of swapping their locations. The three waypoints for each new agent are placed in a similar way as for the existing agents. The upper bound on makespan is 23.

For each query type, we have generated all possible query instances over all agents, waypoints, time steps, etc. For instance, query QC1 asks why an Agent {\small\tt a} charge at location {\small\tt x}. According to the solution for the first \mmapf instance, Agent A1 charges once and Agent A2 charges twice in their plans so we consider all 3 query QC1 instances.  Query QP1 asks why Agent {\small\tt a} does not have a plan whose length is less than {\small\tt l}, so we consider 2 query QP1 instances for each agent with length {\small\tt l}.

\begin{table}[t]
	\caption{\footnotesize Experimental results for (left-table) the \mmapf instance M1 shown in Figure~\ref{fig:fig_qc4} with 2 agents where the upper bound on makespan is 18, and (right-table) the revised \mmapf instance, M2, with 2 more agents located at the empty corners with the goals of swapping their locations and where the upper bound on makespan is 23. The total number of \clingo calls and the answer sets computed in these calls, and the average CPU time (per query instance) in seconds are reported.}
	\label{tab:small}
%\vspace{-\baselineskip}
\begin{tabular}{ll}
	\begin{minipage}{0.5\textwidth}
		\resizebox{0.9\textwidth}{!}{\begin{tabular}{lccc}
			\hline\hline
			Query & \#Instances & \#Calls {[}\#Models{]} & Time (sec) \\
			\hline
	%		QW1 & NA  \\
%			QW2 &  NA \\
%			QW3 & NA \\
%			QW4 & NA \\
			QC1 & 3  & 5 {[}48{]}   & 0.402 \\
			QC2 & 3  & 5 {[}63{]}   & 0.319 \\
			QC3 & 3  & 5 {[}63{]}   & 0.333 \\
			QC4 & 2  & 4 {[}91{]}   & 0.478 \\
			QP1 & 2  & 4 {[}42{]}   & 0.321 \\
			QP2 & 30 & 57 {[}867{]} & 0.394 \\
			QP3 & 30 & 45 {[}545{]} & 0.280  \\
			QP4 & 30 & 53 {[}846{]} & 0.410  \\
			QP5 & 30 & 42 {[}452{]} & 0.241
		\end{tabular}}
	\end{minipage}
&
	\begin{minipage}{0.5\textwidth}
		\resizebox{0.9\textwidth}{!}{\begin{tabular}{lccc}
			\hline\hline
			Query & \#Instances & \#Calls {[}\#Models{]} & Time (sec)\\
			\hline
			QW1 & 4 & 4 [67] & 3.695 \\
			QW2 & 4 & 4 [52] & 3.457 \\
			QW3 & 4 & 4 [52] & 3.605 \\
			QW4 & 4 & 4 [52] & 3.571 \\
			QC1 & 6 & 8 [321] & 45.762 \\
			QC2 & 7 & 7 [87] & 3.381 \\
			QC3 & 7 & 7 [74] & 3.387 \\
			QC4 & 4 & 7 [541] & 100.269 \\
			QP1 & 4 & 6 [306] & 13.806 \\
			QP2 & 59 & 95 [5843] & 27.295 \\
			QP3 & 70 & 70 [898] & 3.425 \\
			QP4 & 66 & 87 [4463] & 18.743 \\
			QP5 & 66 & 66 [852] & 4.127
		\end{tabular}}
	\end{minipage}
\end{tabular}
%\vspace{-2\baselineskip}
\end{table}

\begin{table}[t]
	\caption{\footnotesize  Experimental results for the \mmapf instance M3 with 2 agents, shown in Figure~\ref{fig:fig_qc4}, (left-table) with optimization and (right-table) with anytime search with a time limit of 100 seconds and upper bound for makespan is 45. The total number of \clingo calls and the answer sets computed in these calls, and the average CPU time (per query instance) in seconds are reported.}
	\label{tab:large}
%\vspace{-\baselineskip}
\begin{tabular}{ll}
	\begin{minipage}{0.5\textwidth}
		\resizebox{0.9\textwidth}{!}{\begin{tabular}{lccc}
			\hline\hline
			Query & \#Instances & \#Calls {[}\#Models{]} & Time (sec) \\
			\hline
		QW1 &  2 & 2 {[}13{]} & 57.652 \\
		QW2 &  2 & 2 {[}17{]} & 57.319 \\
		QW3 &  2 & 2 {[}17{]} & 57.462 \\
		QW4 &  2 & 2 {[}17{]} & 57.674 \\
		QC1 &  6 & 11 {[}363{]} & 2437.347 \\
		QC2 &  8 & 12 {[}266{]} & 428.983 \\
		QC3 &  8 & 12 {[}328{]} & 370.401 \\
		QC4 &  2 & 4 {[}159{]} & 6576.363 \\
		QP1 &  2 & 4 {[}166{]} & 579.143 \\
		QP2 &  63 & 116 {[}4277{]} & 1083.826 \\
		QP3 &  74 & 113 {[}2893{]} & 510.654 \\
		QP4 &  71 & 119 {[}4053{]} & 985.894 \\
		QP5 &  72 & 109 {[}2824{]} & 516.365
		\end{tabular}}
	\end{minipage}
&
	\begin{minipage}{0.5\textwidth}
		\resizebox{0.9\textwidth}{!}{\begin{tabular}{lccc}
			\hline\hline
			Query & \#Instances & \#Calls {[}\#Models{]} & Time (sec)\\
			\hline
			QW1 & 2 & 2 {[}13{]} & 53.282 \\
			QW2 & 2 & 2 {[}17{]} & 48.105 \\
			QW3 & 2 & 2 {[}17{]} & 47.203 \\
			QW4 & 2 & 2 {[}17{]} & 47.265 \\
			QC1 & 6 & 11 {[}254{]} & 125.401 \\
			QC2 & 8 & 12 {[}209{]} & 94.569 \\
			QC3 & 8 & 12 {[}288{]} & 93.489 \\
			QC4 & 2 & 4 {[}89{]} & 181.45 \\
			QP1 & 2 & 4 {[}142{]} & 129.776 \\
			QP2 & 63 & 116 {[}3109{]} & 102.225 \\
			QP3 & 74 & 113 {[}2222{]} & 98.412 \\
			QP4 & 71 & 119 {[}2951{]} & 99.575 \\
			QP5 & 72 & 109 {[}1966{]} & 95.451
		\end{tabular}}
	\end{minipage}
\end{tabular}
%\vspace{-1\baselineskip}
\end{table}

For each query instance, we have run our algorithm to generate an explanation. For each query, we report the average CPU time in seconds, the total number of calls to \clingo and the number of answer sets computed by \clingo within these calls.
The results of these experiments are shown in Tables~\ref{tab:small}(left) and \ref{tab:small}(right). Note that the queries QW1--QW4 are not applicable for the first \mmapf instance since no agent waits.  For example, in Table~\ref{tab:small}(left), for QP2 over the \mmapf instance M1, explanations are generated for 30 query instances. In total, 30 QP2 query instances call \clingo 57 times (30 calls with hard constraints, and 27 calls with relevant weak constraints) where 867 answer sets are computed during optimizations; the average computation time for explanation generation per query instance is 0.394 seconds. For the \mmapf instance M2 (Table~\ref{tab:small}(right)), 59 QP2 instances call \clingo 95 times (59 calls with hard constraints, and 36 calls with relevant weak constraints) where 5843 answer sets are computed during optimizations; the average CPU time for explanation generation per instance is 102.225 seconds.

We can observe from these results that doubling the number of agents increases the computation times. The explanations for most of the query instances over the \mmapf instance M1 include recommendations for alternative plans. For the \mmapf instance M2, only some of query QC1, QC4, QP1, QP2, QP4 instances include recommendations for alternative plans; that is why the CPU times are larger for these queries.

We have also experimented with a \mmapf instance, M3, obtained from the first instance M1 by replicating the environment twice, towards the right side, as illustrated in Figure~\ref{fig:fig_m3}. The two agents are initially placed at the far opposite corners in the same way as in the first instance, with the goal of swapping their locations. The results are shown in Table~\ref{tab:large}(left). We have observed from these results that doubling the size of the environment (and thus the plan) increases the computation times.   For the \mmapf instance M3, 63 QP2 instances call \clingo 116 times (63 calls with hard constraints, and 53 calls with relevant weak constraints) where 4277 answer sets are computed during optimizations; the average CPU time for explanation generation per instance is around 18 minutes. This is not surprising since the increase in grid size has a significant effect in the computation time for MAPF problems, as also observed in our earlier studies~\cite{ErdemKOS13}.

For the \mmapf instance M3, we have also experimented with \clingo by utilizing its anytime search feature with a time threshold of 100 seconds: \clingo reports the best solution it computes within 100 seconds. The results are shown in Table~\ref{tab:large}(right). Then, the average computation times reduce significantly. For instance, for QP2 instances, the average CPU time for generating an explanation reduces to 102.225 seconds.

%----------------------------------------------------
%\vspace{-2ex}
\section{Discussion and Conclusion}

We present a novel method for generating a variety of explanations for \mmapf problems, motivated by real life applications in autonomous warehouses, using answer set programming. These explanations are requested by different types of queries posed by the users interactively. In that sense, it is useful for the users to better understand the strengths and weaknesses of the plans being executed by robots at their warehouses, and the limitations of the warehouse infrastructure.

This contribution of query-based explanation generation is important also from the perspective of studies on \mapf. \mapf has been investigated in AI using search algorithms~\cite{Silver05,LunaB11,DresnerS08,WangB08,JansenS08,ChouhanN15,SharonSFS15,SternSFK0WLA0KB19}
or declarative methods~\cite{YuL13,Surynek12ictai,ErdemKOS13}. However, explainability for \mapf problems, as described above, has not been investigated in the literature. The only relevant work that studies explainability for \mapf problems is very recently published: explainability is understood as verification of whether a given plan involves collisions~\cite{AlmagorL20}, and the authors introduce a decomposition-based search method for such explanation schemes. In that sense, our study is useful for \mapf studies by providing a novel query-based declarative method for generating a variety of knowledge-rich explanations for a general variant of \mapf.

Explainability of plans has been emphasized by~\citeN{Smith12} for planning as an iterative process. As planning domains approximate the real-world and optimization functions may not reflect the desired conditions, the computed plans may not be as good as expected.  Smith suggests, in such cases, that planning should be an iterative process where the users inspect the plans and provide feedback to the planner for further improvement. In this process, Smith points out that providing explanations to following questions plays an important role: Why is a given action included in the plan? Why is this action done before that one? Why does the plan not satisfy this property? Why does the plan not achieve this goal? Note that all these questions can be handled by QP, QW, QC and QU queries in the context of \mmapf. Recently, \citeN{EiflerC0MS20} has presented a method to generate ``contrastive'' explanations to questions of the last two forms: `` Why not~$p$?'' where $p$ is a propositional formula describing an plan property.  The underlying idea is to generate an explanation based on the properties $q$ (selected from a given set of properties) entailed by $p$: ``because that would necessitate $\neg q$.'' The authors, in particular, focus on properties $p$ (called ``action-set'' properties) that are about actions involved in the plan, like ``vertex $x$ is visited by agent $y$'' or ``edge from $x$ to $y$ is used by agent $y$.'' For instance, one of the questions they study in the IPC NoMystery domain---a transportation domain with trucks delivering packages to destinations---is ``Why does truck $T0$ not avoid the road from location $L0$ to location $L5$?''  Note that this question is very similar to the question QP4 stated in Scenario 5. In that sense, our approach of utilizing hard constraints and weak constraints is general enough to generate explanations to questions investigated for plan explainability.

Our method is based on an algorithm that utilizes weighted weak constraints of answer set programming for generating explanations by means of counterfactuals, and that allows a sequence of interactive query answering by means of hypothetical reasoning. This contribution of counterfactual-based explanation generation using weighted weak constraints is important also from the perspective of studies on explanation generation in ASP. Explanation generation has been investigated in answer set programming, based on justifications,  debugging and/or argumentations~\cite{pontelli2009justifications,schulz2013aba,schulz2016justifying,cabalar2014causal,cabalar2016justifications,damasio2013justifications,brain2007illogical,gebser2008meta,oetsch2010catching,ErdemO15}, as summarized in the surveys~\cite{fandinno2018answering,DodaroGRRS19}. For instance, in our earlier studies~\cite{ErdemO15}, we generate explanations for complex biomedical queries for drug discovery (expressed in a controlled natural language), based on the idea of finding justifications.  Our method extends this list by the use of weighted weak constraints.

%Our ongoing work includes extending the variety of types of queries, and investigating other ASP approaches to generate explanations for \mmapf problems.

%----------------------------------------------------

%\vspace{-2ex}
\paragraph{\bf Acknowledgments}
We have benefited from useful discussions with Alexander Kleiner (Robert Bosch GmbH, Corporate Research) and Volkan Patoglu (Sabanci University, Mechatronics Engineering) about real world robotic applications of \mmapf in warehouses.

%----------------------------------------------------

\bibliographystyle{acmtrans}
%\bibliography{references}

\begin{thebibliography}{}

\bibitem[\protect\citeauthoryear{Almagor and Lahijanian}{Almagor and
  Lahijanian}{2020}]{AlmagorL20}
{\sc Almagor, S.} {\sc and} {\sc Lahijanian, M.} 2020.
\newblock Explainable multi agent path finding.
\newblock In {\em Proc. of AAMAS}. 34--42.

\bibitem[\protect\citeauthoryear{Bogatarkan, Erdem, Kleiner, and
  Patoglu}{Bogatarkan et~al\mbox{.}}{2020}]{bogatarkan2020multi}
{\sc Bogatarkan, A.}, {\sc Erdem, E.}, {\sc Kleiner, A.}, {\sc and} {\sc
  Patoglu, V.} 2020.
\newblock Multi-modal multi-agent path finding with optimal resource
  utilization.
\newblock In {\em Proceedings of 5th International Conference on the Industry
  4.0 Model for Advanced Manufacturing}. 313--324.

\bibitem[\protect\citeauthoryear{Bogatarkan, Patoglu, and Erdem}{Bogatarkan
  et~al\mbox{.}}{2019}]{BogatarkanPE19}
{\sc Bogatarkan, A.}, {\sc Patoglu, V.}, {\sc and} {\sc Erdem, E.} 2019.
\newblock A declarative method for dynamic multi-agent path finding.
\newblock In {\em Proceedings of the 5th Global Conference on Artificial
  Intelligence}. 54--67.

\bibitem[\protect\citeauthoryear{Brain, Gebser, Pührer, Schaub, Tompits, and
  Woltran}{Brain et~al\mbox{.}}{2007}]{brain2007illogical}
{\sc Brain, M.}, {\sc Gebser, M.}, {\sc Pührer, J.}, {\sc Schaub, T.}, {\sc
  Tompits, H.}, {\sc and} {\sc Woltran, S.} 2007.
\newblock “that is illogical captain!”–the debugging support tool spock
  for answer-set programs: System description.
%\newblock 71--85.

\bibitem[\protect\citeauthoryear{Cabalar and Fandinno}{Cabalar and
  Fandinno}{2016}]{cabalar2016justifications}
{\sc Cabalar, P.} {\sc and} {\sc Fandinno, J.} 2016.
\newblock Justifications for programs with disjunctive and causal-choice rules.
\newblock {\em Theory Pract. Log. Program.\/}~{\em 16,\/}~5-6, 587--603.

\bibitem[\protect\citeauthoryear{Cabalar, Fandinno, and Fink}{Cabalar
  et~al\mbox{.}}{2014}]{cabalar2014causal}
{\sc Cabalar, P.}, {\sc Fandinno, J.}, {\sc and} {\sc Fink, M.} 2014.
\newblock Causal graph justifications of logic programs.
\newblock {\em Theory Pract. Log. Program.\/}~{\em 14,\/}~4-5, 603--618.

\bibitem[\protect\citeauthoryear{Calimeri, Faber, Gebser, Ianni, Kaminski,
  Krennwallner, Leone, Ricca, and Schaub}{Calimeri
  et~al\mbox{.}}{2020}]{aspcore2}
{\sc Calimeri, F.}, {\sc Faber, W.}, {\sc Gebser, M.}, {\sc Ianni, G.}, {\sc
  Kaminski, R.}, {\sc Krennwallner, T.}, {\sc Leone, N.}, {\sc Ricca, F.}, {\sc
  and} {\sc Schaub, T.} 2020.
\newblock {ASP-Core-2} Input Language Format.
\newblock {\em Theory Pract. Log. Program.\/}~{\em 20,\/}~2, 294--309.

\bibitem[\protect\citeauthoryear{Chouhan and Niyogi}{Chouhan and
  Niyogi}{2015}]{ChouhanN15}
{\sc Chouhan, S.~S.} {\sc and} {\sc Niyogi, R.} 2015.
\newblock {DMAPP:} {A} distributed multi-agent path planning algorithm.
\newblock In {\em Proc. of AI}. 123--135.

%\bibitem[\protect\citeauthoryear{Cyras, Letsios, Misener, and Toni}{Cyras
%  et~al\mbox{.}}{2019}]{CyrasLMT19}
%{\sc Cyras, K.}, {\sc Letsios, D.}, {\sc Misener, R.}, {\sc and} {\sc Toni, F.}
%  2019.
%\newblock Argumentation for explainable scheduling.
%\newblock In {\em Proc. of AAAI}. 2752--2759.

\bibitem[\protect\citeauthoryear{Dam{\'{a}}sio, Analyti, and
  Antoniou}{Dam{\'{a}}sio et~al\mbox{.}}{2013}]{damasio2013justifications}
{\sc Dam{\'{a}}sio, C.~V.}, {\sc Analyti, A.}, {\sc and} {\sc Antoniou, G.}
  2013.
\newblock Justifications for logic programming.
\newblock In {\em Proc. of LPNMR}. 530--542.

\bibitem[\protect\citeauthoryear{Dijkstra}{Dijkstra}{1959}]{Dijkstra59}
{\sc Dijkstra, E.~W.} 1959.
\newblock A note on two problems in connexion with graphs.
\newblock {\em Numer. Math.\/}~{\em 1,\/}~1, 269--271.

\bibitem[\protect\citeauthoryear{Dodaro, Gasteiger, Reale, Ricca, and
  Schekotihin}{Dodaro et~al\mbox{.}}{2019}]{DodaroGRRS19}
{\sc Dodaro, C.}, {\sc Gasteiger, P.}, {\sc Reale, K.}, {\sc Ricca, F.}, {\sc
  and} {\sc Schekotihin, K.} 2019.
\newblock Debugging non-ground {ASP} programs: Technique and graphical tools.
\newblock {\em Theory Pract. Log. Program.\/}~{\em 19,\/}~2, 290--316.

\bibitem[\protect\citeauthoryear{Dresner and Stone}{Dresner and
  Stone}{2008}]{DresnerS08}
{\sc Dresner, K.~M.} {\sc and} {\sc Stone, P.} 2008.
\newblock A multiagent approach to autonomous intersection management.
\newblock {\em J. Artif. Intell. Res. (JAIR)\/}~{\em 31}, 591--695.

\bibitem[\protect\citeauthoryear{Eifler, Cashmore, Hoffmann, Magazzeni, and
  Steinmetz}{Eifler et~al\mbox{.}}{2020}]{EiflerC0MS20}
{\sc Eifler, R.}, {\sc Cashmore, M.}, {\sc Hoffmann, J.}, {\sc Magazzeni, D.},
  {\sc and} {\sc Steinmetz, M.} 2020.
\newblock A new approach to plan-space explanation: Analyzing plan-property
  dependencies in oversubscription planning.
\newblock In {\em Proc. of AAAI}. 9818--9826.

\bibitem[\protect\citeauthoryear{Erdem, Kisa, Oztok, and Schueller}{Erdem
  et~al\mbox{.}}{2013}]{ErdemKOS13}
{\sc Erdem, E.}, {\sc Kisa, D.~G.}, {\sc Oztok, U.}, {\sc and} {\sc Schueller,
  P.} 2013.
\newblock A general formal framework for pathfinding problems with multiple
  agents.
\newblock In {\em Proc. of AAAI}.

\bibitem[\protect\citeauthoryear{Erdem and {\"{O}}ztok}{Erdem and
  {\"{O}}ztok}{2015}]{ErdemO15}
{\sc Erdem, E.} {\sc and} {\sc {\"{O}}ztok, U.} 2015.
\newblock Generating explanations for biomedical queries.
\newblock {\em Theory Pract. Log. Program.\/}~{\em 15,\/}~1, 35--78.

\bibitem[\protect\citeauthoryear{Fandinno and Schulz}{Fandinno and
  Schulz}{2019}]{fandinno2018answering}
{\sc Fandinno, J.} {\sc and} {\sc Schulz, C.} 2019.
\newblock Answering the "why" in answer set programming - {A} survey of
  explanation approaches.
\newblock {\em Theory Pract. Log. Program.\/}~{\em 19,\/}~2, 114--203.

\bibitem[\protect\citeauthoryear{Gebser, P{\"{u}}hrer, Schaub, and
  Tompits}{Gebser et~al\mbox{.}}{2008}]{gebser2008meta}
{\sc Gebser, M.}, {\sc P{\"{u}}hrer, J.}, {\sc Schaub, T.}, {\sc and} {\sc
  Tompits, H.} 2008.
\newblock A meta-programming technique for debugging answer-set programs.
\newblock In {\em Proc. of AAAI}. 448--453.

\bibitem[\protect\citeauthoryear{Gelfond and Lifschitz}{Gelfond and
  Lifschitz}{1988}]{gelfond1988stable}
{\sc Gelfond, M.} {\sc and} {\sc Lifschitz, V.} 1988.
\newblock The stable model semantics for logic programming.
\newblock In {\em Proceedings of International Logic Programming Conference and
  Symposium}. 1070--1080.

\bibitem[\protect\citeauthoryear{Gelfond and Lifschitz}{Gelfond and
  Lifschitz}{1991}]{gelfondL91}
{\sc Gelfond, M.} {\sc and} {\sc Lifschitz, V.} 1991.
\newblock Classical negation in logic programs and disjunctive databases.
\newblock {\em New Generation Computing\/}~{\em 9}, 365--385.

\bibitem[\protect\citeauthoryear{Jansen and Sturtevant}{Jansen and
  Sturtevant}{2008}]{JansenS08}
{\sc Jansen, R.} {\sc and} {\sc Sturtevant, N.} 2008.
\newblock A new approach to cooperative pathfinding.
\newblock In {\em Proc. of AAMAS}. 1401--1404.

\bibitem[\protect\citeauthoryear{Lifschitz}{Lifschitz}{2002}]{Lifschitz02}
{\sc Lifschitz, V.} 2002.
\newblock Answer set programming and plan generation.
\newblock {\em Artificial Intelligence\/}~{\em 138}, 39--54.

\bibitem[\protect\citeauthoryear{Luna and Bekris}{Luna and
  Bekris}{2011}]{LunaB11}
{\sc Luna, R.} {\sc and} {\sc Bekris, K.~E.} 2011.
\newblock Efficient and complete centralized multi-robot path planning.
\newblock In {\em Proc. of IROS}. 3268--3275.

\bibitem[\protect\citeauthoryear{Marek and Truszczy\'nski}{Marek and
  Truszczy\'nski}{1999}]{MarekT99}
{\sc Marek, V.} {\sc and} {\sc Truszczy\'nski, M.} 1999.
\newblock Stable models and an alternative logic programming paradigm.
\newblock In {\em The Logic Programming Paradigm: a 25-Year Perspective}.
  Springer Verlag, 375--398.

\bibitem[\protect\citeauthoryear{Niemel{\"a}}{Niemel{\"a}}{1999}]{Niemelae99}
{\sc Niemel{\"a}, I.} 1999.
\newblock Logic programs with stable model semantics as a constraint
  programming paradigm.
\newblock {\em Annals of Mathematics and Artificial Intelligence\/}~{\em 25},
  241--273.

\bibitem[\protect\citeauthoryear{Oetsch, P{\"{u}}hrer, and Tompits}{Oetsch
  et~al\mbox{.}}{2010}]{oetsch2010catching}
{\sc Oetsch, J.}, {\sc P{\"{u}}hrer, J.}, {\sc and} {\sc Tompits, H.} 2010.
\newblock Catching the ouroboros: On debugging non-ground answer-set programs.
\newblock {\em Theory Pract. Log. Program.\/}~{\em 10,\/}~4-6, 513--529.

\bibitem[\protect\citeauthoryear{Pontelli, Son, and El{-}Khatib}{Pontelli
  et~al\mbox{.}}{2009}]{pontelli2009justifications}
{\sc Pontelli, E.}, {\sc Son, T.~C.}, {\sc and} {\sc El{-}Khatib, O.} 2009.
\newblock Justifications for logic programs under answer set semantics.
\newblock {\em Theory Pract. Log. Program.\/}~{\em 9,\/}~1, 1--56.

\bibitem[\protect\citeauthoryear{Ratner and Warmuth}{Ratner and
  Warmuth}{1986}]{RatnerW86}
{\sc Ratner, D.} {\sc and} {\sc Warmuth, M.~K.} 1986.
\newblock Finding a shortest solution for the n $\times$ n extension of the
  15-puzzle is intractable.
\newblock In {\em Proc. of AAAI}. 168--172.

\bibitem[\protect\citeauthoryear{Schulz and Toni}{Schulz and
  Toni}{2013}]{schulz2013aba}
{\sc Schulz, C.} {\sc and} {\sc Toni, F.} 2013.
\newblock Aba-based answer set justification.
\newblock {\em Theory Pract. Log. Program.\/}~{\em
  13,\/}~4-5-Online-Supplement.

\bibitem[\protect\citeauthoryear{Schulz and Toni}{Schulz and
  Toni}{2016}]{schulz2016justifying}
{\sc Schulz, C.} {\sc and} {\sc Toni, F.} 2016.
\newblock Justifying answer sets using argumentation.
\newblock {\em Theory Pract. Log. Program.\/}~{\em 16,\/}~1, 59--110.

\bibitem[\protect\citeauthoryear{Sharon, Stern, Felner, and Sturtevant}{Sharon
  et~al\mbox{.}}{2015}]{SharonSFS15}
{\sc Sharon, G.}, {\sc Stern, R.}, {\sc Felner, A.}, {\sc and} {\sc Sturtevant,
  N.~R.} 2015.
\newblock Conflict-based search for optimal multi-agent pathfinding.
\newblock {\em Artif. Intell.\/}~{\em 219}, 40--66.

\bibitem[\protect\citeauthoryear{Silver}{Silver}{2005}]{Silver05}
{\sc Silver, D.} 2005.
\newblock Cooperative pathfinding.
\newblock In {\em Proc. of AIIDE}. 117--122.

\bibitem[\protect\citeauthoryear{Smith}{Smith}{2012}]{Smith12}
{\sc Smith, D.~E.} 2012.
\newblock Planning as an iterative process.
\newblock In {\em Proc. of AAAI}, {J.~Hoffmann} {and} {B.~Selman}, Eds.

\bibitem[\protect\citeauthoryear{Stern, Sturtevant, Felner, Koenig, Ma, Walker,
  Li, Atzmon, Cohen, Kumar, Bart{\'{a}}k, and Boyarski}{Stern
  et~al\mbox{.}}{2019}]{SternSFK0WLA0KB19}
{\sc Stern, R.}, {\sc Sturtevant, N.~R.}, {\sc Felner, A.}, {\sc Koenig, S.},
  {\sc Ma, H.}, {\sc Walker, T.~T.}, {\sc Li, J.}, {\sc Atzmon, D.}, {\sc
  Cohen, L.}, {\sc Kumar, T. K.~S.}, {\sc Bart{\'{a}}k, R.}, {\sc and} {\sc
  Boyarski, E.} 2019.
\newblock Multi-agent pathfinding: Definitions, variants, and benchmarks.
\newblock In {\em Proc. of SOCS}. 151--159.

\bibitem[\protect\citeauthoryear{Surynek}{Surynek}{2012}]{Surynek12ictai}
{\sc Surynek, P.} 2012.
\newblock On propositional encodings of cooperative path-finding.
\newblock In {\em Proc. of ICTAI}. 524--531.

\bibitem[\protect\citeauthoryear{Wang and Botea}{Wang and
  Botea}{2008}]{WangB08}
{\sc Wang, K.-H.~C.} {\sc and} {\sc Botea, A.} 2008.
\newblock Fast and memory-efficient multi-agent pathfinding.
\newblock In {\em Proc. of ICAPS}. 380--387.

\bibitem[\protect\citeauthoryear{Yu and LaValle}{Yu and LaValle}{2013}]{YuL13}
{\sc Yu, J.} {\sc and} {\sc LaValle, S.~M.} 2013.
\newblock Planning optimal paths for multiple robots on graphs.
\newblock In {\em Proc. of ICRA}. 3612--3617.

\end{thebibliography}
%\vspace{-3ex}

\end{document}